\documentclass{article} 
\usepackage{iclr2024_conference,times}

\usepackage{hyperref}
\usepackage{url}
\usepackage{amsmath}
\usepackage{algorithm}
\usepackage{algorithmicx}
\usepackage{algpseudocode}
\usepackage{xspace}
\usepackage{mathrsfs}
\usepackage{multirow}
\usepackage{enumitem}
\usepackage{dutchcal}
\usepackage{graphicx} 
\usepackage{subfigure}
\usepackage{pifont}
\usepackage{svg}
\usepackage{threeparttable}
\usepackage{color}
\usepackage{booktabs}
\usepackage{diagbox}
\usepackage{balance} 
\usepackage{amssymb}
\usepackage{bm}
\usepackage{caption}

\newcommand{\algorithmicinput}{\textbf{Input:}}
\newcommand{\INPUT}{\item[\algorithmicinput]}

\title{PTaRL: Prototype-based Tabular Representation Learning via Space Calibration}

\author{Hangting Ye\textsuperscript{\normalfont 1}, Wei Fan\textsuperscript{\normalfont 2}, Xiaozhuang Song\textsuperscript{\normalfont 3}, Shun Zheng\textsuperscript{\normalfont 4}, He Zhao\textsuperscript{\normalfont 5}, Dandan Guo\textsuperscript{\normalfont 1}\footnotemark[1]\ \ , Yi Chang\textsuperscript{\normalfont 1}\thanks{Corresponding authors.}\\
School of Artificial Intelligence, Jilin University\textsuperscript{\normalfont 1}  University of Oxford\textsuperscript{\normalfont 2}\\
The Chinese University of Hong Kong, Shenzhen\textsuperscript{\normalfont 3}
Microsoft Research\textsuperscript{\normalfont 4}
CSIRO’s Data61\textsuperscript{\normalfont 5}\\
\texttt{yeht2118@mails.jlu.edu.cn},
\texttt{wei.fan@wrh.ox.ac.uk},\\
\texttt{xiaozhuangsong1@link.cuhk.edu.cn}, 
\texttt{shun.zheng@microsoft.com},\\
\texttt{he.zhao@ieee.org},
\texttt{\{guodandan,yichang\}@jlu.edu.cn}
}

\newcommand{\method}{{\textsc{PTaRL}}\xspace}

\iclrfinalcopy 
\begin{document}


\maketitle

\begin{abstract}
Tabular data have been playing a mostly important role in diverse real-world fields, such as healthcare, engineering, finance, etc.
With the recent success of deep learning, many tabular machine learning (ML) methods based on deep networks (e.g., Transformer, ResNet) have achieved competitive performance on tabular benchmarks. However, existing deep tabular ML methods suffer from the \textit{representation entanglement} and \textit{localization}, which largely hinders their prediction performance and leads to  \textit{performance inconsistency} on tabular tasks.
To overcome these problems, we explore a novel direction of \textit{applying prototype learning for tabular ML} and propose a prototype-based tabular representation learning framework, \textsc{PTaRL}, for tabular prediction tasks. The core idea of \textsc{PTaRL} is to construct prototype-based projection space (P-Space) and learn the disentangled representation around global data prototypes. Specifically, \textsc{PTaRL} mainly involves two stages: (i) Prototype Generation, that constructs global prototypes as the basis vectors of P-Space for representation, and (ii) Prototype Projection, that projects the data samples into P-Space and keeps the core global data information via Optimal Transport. Then, to further acquire the disentangled representations, we constrain \textsc{PTaRL} with two strategies: (i) to diversify the coordinates towards global prototypes of different representations within P-Space, we bring up a diversification constraint for representation calibration; (ii) to avoid prototype entanglement in P-Space, we introduce a matrix orthogonalization constraint to ensure the independence of global prototypes. 
Finally, we conduct extensive experiments in \textsc{PTaRL} coupled with state-of-the-art deep tabular ML models on various tabular benchmarks and the results have shown our consistent superiority. 


\end{abstract}

\section{Introduction}
\label{sec:introduction}

Tabular data, usually represented as tables in a relational database with rows standing for the data samples and columns standing for the feature variables (e.g., categorical and numerical features), has been playing a more and more vital role across diverse real-world fields, including healthcare~{\citep{hernandez2022synthetic}, engineering~\citep{ye2023uadb}, advertising~\citep{frosch2010decade}, finance~\citep{assefa2020generating}, etc.
Starting from traditional machine learning methods (e.g., linear regression~\citep{su2012linear}, logistic regression~\citep{wright1995logistic}) to tree-based methods (e.g. XGBoost~\citep{chen2016xgboost}, LightGBM~\citep{ke2017lightgbm}), tabular machine learning has received broad attention from researchers for many decades.

More recently, with the great success of deep networks in computer vision (CV)~\citep{he2016deep} and natural language processing (NLP)~\citep{devlin2018bert}, numerous methods based on deep learning have been proposed for tabular machine learning (ML) to accomplish tabular prediction tasks~\citep{song2019autoint, huang2020tabtransformer, gorishniy2021revisiting, wang2021dcn, chen2022danets, xu2024bishop}. 
For example, ~\citep{song2019autoint} proposed AutoInt based on transformers, ~\citep{gorishniy2021revisiting} further improved AutoInt through better token embeddings, and ~\citep{wang2021dcn} proposed DCN V2 that consists of an MLP-like module and a feature crossing module.


While the recent deep learning solutions have performed competitively on tabular benchmarks~\citep{gorishniy2021revisiting, shwartz2022tabular}, there still exists an \textit{performance inconsistency} in their predictions on tabular tasks:  existing state-of-the-art deep tabular ML models (e.g., FT-Transformer~\citep{gorishniy2021revisiting}, ResNet~\citep{gorishniy2021revisiting}) cannot perform consistently well on different tasks, such as regression, classification, etc. 
We investigate the learned patterns of deep tabular ML models and identify two inherent characteristic hindering prediction: 
(i) \textit{representation entanglement}: the learned representations of existing deep tabular methods are usually entangled and thus cannot support clear and accurate decision-making, and (ii) \textit{representation localization}: each data sample are represented distinctively, making the global data structures over data samples are overlooked.

To better overcome the aforementioned challenges, we explore the direction of \textit{applying prototype learning for tabular modeling}, and accordingly we propose \textsc{PTaRL}, a prototype-based tabular representation learning framework for tabular ML predictions. The core idea of \textsc{PTaRL} is to develop a \textit{prototype-based projection space} (P-Space) for deep tabular ML models, in which the disentangled representations\footnote{In our paper, ``disentangled representations'' means the representations are more separated and discriminative for supervised tabular modeling tasks, which is different from disentanglement in deep generative models.} around pre-defined global prototypes can be acquired with global tabular data structure to enhance the tabular predictions.
Specifically, our \textsc{PTaRL} mainly involves two stages, (i) Prototype Generating and (ii) Representation Projecting.
In the first stage, we construct $K$ global prototypes for tabular representations, each of which is regarded as the basis vector for the P-Space to stimulate disentangled learning for more global data representations. We initialize the global prototypes with \textit{K-Means clustering}~\citep{hartigan1979algorithm} to facilitate the efficiency of prototype learning. 
In the second stage, we project the original data samples into P-Space with the global prototypes to learn the representations with global data structure information.
To learn the global data structure, we propose a shared estimator to output the projected representations with global prototypes; besides, we propose to utilize \textit{Optimal Transport}~\citep{peyre2017computational} to jointly optimize the learned representations in P-Space with global prototypes and original representations by deep tabular models, to preserve original data structure information.


In addition to employing global prototypes, we propose two additional strategies to further disentangle the learned representations in \textsc{PTaRL}:
(i) Coordinates Diversifying Constraint motivated by contrastive learning that diversifies the representation coordinates of data samples in P-Space to represent data samples in a disentangled manner, and (ii) Matrix Orthogonalization Constraint that makes the defined global prototypes in P-Space orthogonal with each other to ensure the independence of prototypes and facilitate the disentangled learning. In brief, our contribution can be summarized as follows:
\begin{itemize}[leftmargin=*]
    \item We investigated the learned patterns of deep tabular models and explore a novel direction of \textit{applying prototype learning for tabular machine learning} to address representation entanglement and localization.
    \item We propose a model-agnostic prototype-based tabular representation learning framework, 
    \textsc{PTaRL} for tabular prediction tasks, which transforms data into the prototype-based projection space and optimize representations via Optimal Transport.
    \item We propose two different strategies, the Coordinates Diversifying Constraint and the Matrix Orthogonalization Constraint to make \textsc{PTaRL} learn disentangled representations.
    \item We conducted extensive experiments in \textsc{PTaRL} coupled with state-of-the-art (SOTA) deep tabular ML models on various tabular benchmarks and the comprehensive results along with analysis and visualizations demonstrate our effectiveness.
\end{itemize}

\section{Related Work}

\textbf{Deep Learning for Tabular machine learning.}
Starting from statistical machine learning methods (e.g., linear regression~\citep{su2012linear}, logistic regression~\citep{wright1995logistic}) to tree-based methods (e.g. XGBoost~\citep{chen2016xgboost}, LightGBM~\citep{ke2017lightgbm}), traditional machine learning methods are broadly used for tabular machine learning. 
More recently, inspired by the success of of deep learning in computer vision (CV)~\citep{he2016deep} and natural language processing (NLP)~\citep{devlin2018bert}, numerous methods based on deep learning have been proposed for tabular machine learning to accomplish tabular prediction tasks~\citep{song2019autoint, huang2020tabtransformer, gorishniy2021revisiting, wang2021dcn}. 
Among these works, ~\citet{wang2021dcn} proposed DCN V2 that consists of an MLP-like module and a feature crossing module; AutoInt~\citep{song2019autoint} leveraged the Transformer architecture to capture inter-column correlations; FT-Transformer~\citep{gorishniy2021revisiting} further enhanced AutoInt's performance through improved token embeddings; ResNet for tabular domain ~\citep{gorishniy2021revisiting} also achieved remarkable performance.
However, these methods may fail to capture the global data structure information, and are possibly affected by the representation coupling problem. Therefore, they cannot perform consistently well on different tasks, e.g. regression and classification.
Recently, another line of research has tried to use additional information outside target dataset to enhancing deep learning for tabular data.
TransTab ~\citep{wang2022transtab} incorporates feature name information into Transformer to achieve cross table learning.
XTab ~\citep{zhu2023xtab} pretrains Transformer on a variety of datasets to enhance tabular deep learning. 
Different from this line, \method does not need additional information outside target dataset.
Note that \method, as a general representation learning pipeline, is model-agnostic such that it can be integrated with many of the above deep tabular ML models to learn better tabular data representations.

\paragraph{Prototype Learning.}
Typically, a prototype refers to a proxy of a class and it is computed as the weighted results of latent features of all instances belonged to the corresponding class. Prototype-based methods have been widely applied in a range of machine learning applicaitons, like computer vision~\citep{yang2018robust,li2021adaptive,nauta2021neural,zhou2022rethinking}, natural language processing~\citep{huang2012improving,  devlin2018bert,zalmout2022prototype}. In the field of CV, prototype learning assigns labels to testing images by computing their distances to prototypes of each class. This method has been proven to make model to be more resilient and consistent when dealing with few-shot and zero-shot scenarios~\citep{yang2018robust}. Likewise, in the field of natural language processing (NLP), taking the mean of word embeddings as prototypes for sentence representations has also demonstrated robust and competitive performance on various NLP tasks.


The mentioned approachs generally employ the design of prototype learning to facilitate the sharing of global information across tasks, enabling rapid adaptation of new tasks~\citep{huang2012improving, hoang20b, li2021adaptive,zhou2022rethinking}. Similarly, in tabular deep learning, the global information of data samples is crucial for inferring labels of each data sample~\citep{zhou2020table2analysis,du2021tabularnet}. This inspired us to incorporate prototype learning into our proposed framework for capturing global information and leveraging it to enhance the tabular learning performance.

%


\section{Background}
\label{sec: background}
\textbf{Notation.} 
Denote the $i$-th sample as $(x_i, y_i)$,  where $x_i = (x_i^{(num)}, x_i^{(cat)}) \in \mathbb{X}$ represents numerical and categorical features respectively and $y_i \in \mathbb{Y}$ is the corresponding label.
A tabular dataset $D=\{X, Y\}$ is a collection of $n$ data samples, where $X=\{x_i\}_{i=1}^n$ and $Y=\{y_i\}_{i=1}^n$. 
We use $D_{train}$ to denote training set for training, $D_{val}$ to denote validation set for early stopping and hyperparameter tuning, and $D_{test}$ to denote test set for final evaluation.
Note that in this paper we consider deep learning for supervised tabular prediction tasks: binary classification $\mathbb{Y} = \{0, 1\}$, multiclass classification $\mathbb{Y} = \{1, . . . , c\}$ and regression $\mathbb{Y} = \mathbb{R}$. 
The goal is to obtain an accurate deep tabular model $F(\cdot;\theta): \mathbb{X} \to \mathbb{Y}$ trained on $D_{train}$.

\textbf{Optimal Transport.}
Although Optimal Transport (OT) possesses a rich theoretical foundation, we focus our discussion solely on OT for discrete probability distributions, please refer to ~\citep{peyre2017computational} for more details. Let us consider $p$ and $q$ as two discrete probability distributions over an arbitrary space $\mathbb{S} \in \mathbb{R}^d$, which can be expressed as $p=\sum_{i=1}^{n} a_i \delta_{x_i}$ and $q=\sum_{j=1}^{m} b_j \delta_{y_j}$. 
In this case, $\bm{a} \in \sum^{n}$ and $\bm{b} \in \sum^{m}$, where $\sum^{n}$ represents the probability simplex in $\mathbb{R}^n$. The OT distance between $p$ and $q$ is defined as:
\begin{equation}
\label{ot_def}
\begin{aligned}
    \text{OT}(p, q) = \min_{\textbf{T}\in \Pi (p, q)} \langle \textbf{T}, \textbf{C} \rangle ,
\end{aligned}
\end{equation}
 where $\langle \cdot, \cdot \rangle$  is the Frobenius dot-product and $\textbf{C} \in \mathbb{R}^{n\times m}_{\geq 0}$ is the transport cost matrix constructed by $C_{ij} = Distance(x_i, y_j)$.
The transport probability matrix $\textbf{T} \in \mathbb{R}^{n\times m}_{\geq 0}$, which satisfies $\Pi(p,q):= \{\textbf{T} | \sum_{i=1}^{n} T_{ij}=b_j,  \sum_{j=1}^{m} T_{ij}=a_i\}$, is learned by minimizing $\langle \textbf{T}, \textbf{C} \rangle$. 
Directly optimizing Eq.~\ref{ot_def} often comes at the cost of heavy computational demands, and OT with entropic regularization is
introduced to allow the optimization at small computational cost in sufficient smoothness ~\citep{cuturi2013sinkhorn}.

\section{Proposed Method: \method}

\begin{figure}[h]
\centering
\includegraphics[width=\linewidth]{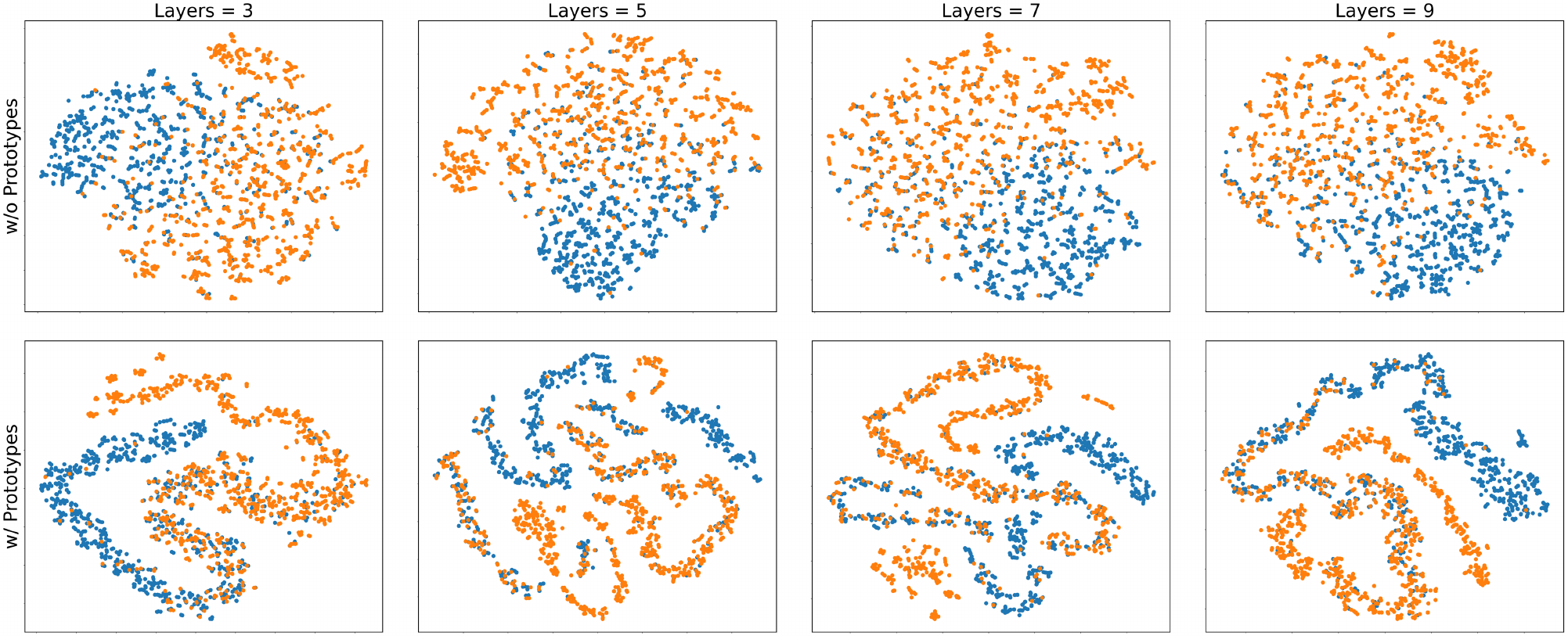}
\captionsetup{font=small}
\caption{The visualization of representations of deep network w/o and w/ \method with varying model layer depths.
}
\label{fig:source}
\vspace{-1em}
\end{figure}

\subsection{Motivation and Overall Pipeline}
\label{sec:entanglement details} 
In the context of tabular data, the intrinsic heterogeneity presents a challenge for achieving satisfactory performance using deep models. 
As shown in the first subfigure of Fig.~\ref{fig:source}, the latent representations learned by the FT-Transformer ~\citep{gorishniy2021revisiting} (one of the SOTA deep tabular models) on a binary classification dataset Adult ~\citep{kohavi1996scaling} are entangled.
To validate whether it is caused by the limitation of model capacity, we gradually increase FT-Transformer's layer depths and visualize the corresponding latent representation by T-SNE ~\citep{van2008visualizing}. As shown in the first row of Fig.~\ref{fig:source}, with a sequential increase in the number of model layers, we could observe the \textit{representation entanglement} phenomenon has not be mitigated while gradually augmenting the model capacity.
In addition, our empirical observations indicate that this phenomenon also exists for other deep tabular models and we recognize that the representation entanglement is the inherent limitation of deep models in tabular domain.
Moreover, the learned representations also lack global data structure information, which failed to model the shared information among the data instances.
Compared to other domains like CV and NLP, especially in the heterogeneous tabular domain, samples overlook the statistical global structure information among the total dataset would drop into \textit{representation localization}. 
Furthermore, recent researches ~\citep{gorishniy2021revisiting, shwartz2022tabular} show that different types of data may require varying types of deep models (e.g. Transformer based and MLP based architecture).

To address the aforementioned limitations of deep models for the tabular domain, we apply prototype learning into tabular modeling and propose the prototype-based tabular representation learning (\method) framework. Note that \method, as a general representation learning framework, is model-agnostic such that it can be coupled with any deep tabular model $F(\cdot;\theta)$ to learn better tabular data representations in our redefined \textit{Prototype-based Projection Space}, which is the core of \method.
In the following, we will elaborate on the specific learning procedure of our \method in Section ~\ref{sec:ptarl_main}; then, we will provide two constraints to further constraint \method for representation calibration and prototype independence in Section \ref{sec:ptarl_constraints}. 
As shown in the second row of Fig.~\ref{fig:source}, with \method, the latent space is calibrated to make the representation disentangled.
Fig.~\ref{fig:tab_framework} gives an overview of the proposed \method.
Before the introduction of \method, we first present the formal definition of the prototype-based projection space as follows:

\textbf{Definition 1. Prototype-based Projection Space (P-Space).} Given a set of global prototypes $\mathcal{B} = \{\beta_k\}_{k=1}^K \in \mathbb{R}^{K\times d}$ where $K$ is the number of prototypes and $\beta_k$ is the representation of the $k$-th prototype with $d$ as the hidden dimension, we define the P-Space as a projection space consisting of the global prototypes $\mathcal{B}$ as the basis vectors for representation.
For example, given a representation in P-Space with \textbf{\textit{coordinates}} denoted as $r = \{r^k\}_{k=1}^{K} \in \mathbb{R}^{K\times 1}$, each representation in P-Space can be formulated as $\sum_{k=1}^{K} r^k \beta_k$.

\begin{figure}[!t]
\centering
\includegraphics[width=\linewidth]{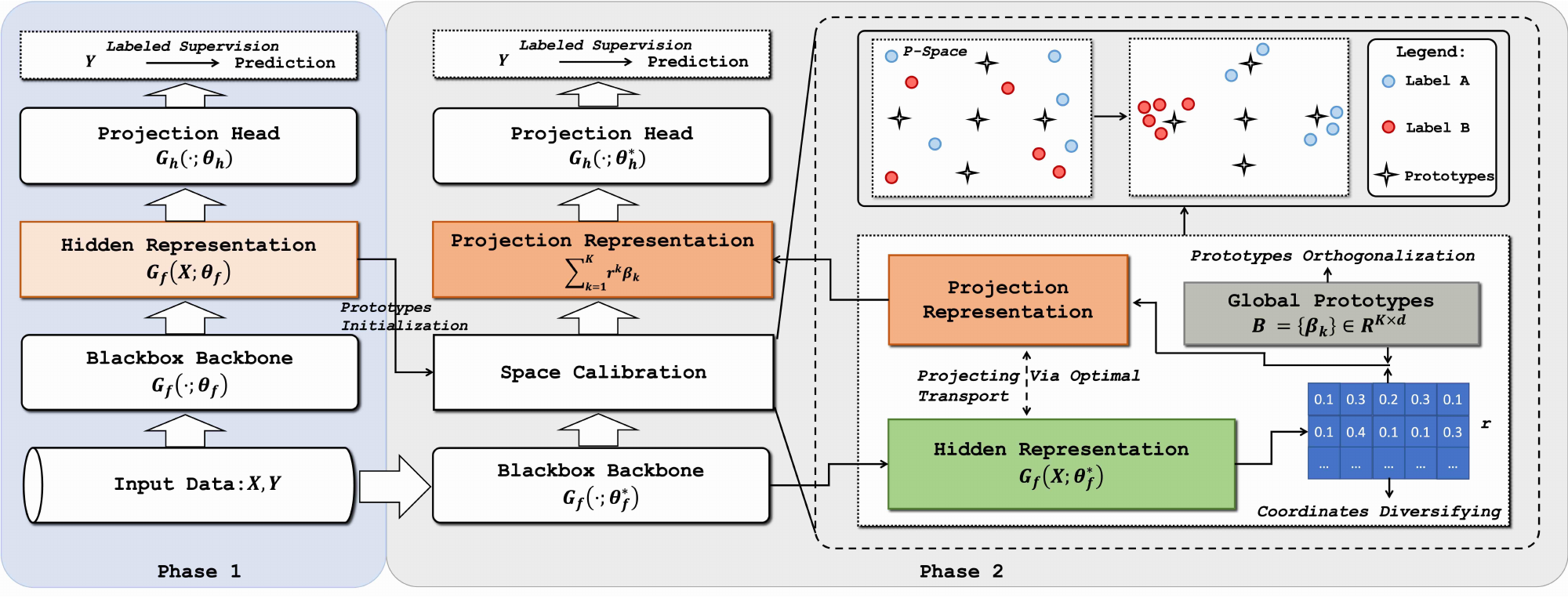}
\captionsetup{font=small}
\captionsetup{skip=0pt}
\caption{The \method framework. The original representation of each sample by backbone would be pushed forward to the corresponding projection representation by minimizing the Optimal Transport Distance. The two sentences ``coordinates diversifying'' and ``prototypes orthogonalization'' correspond to two constraints for representation disentanglement.} 


\label{fig:tab_framework}
\vspace{-1em}
\end{figure}

\subsection{Prototype-based Tabular Representation Learning (\method)} \label{sec:ptarl_main}
\method consists of the following two stages: (i) prototype generating, that constructs global prototypes of P-Space, and (ii) representation projecting, that projects $F(\cdot;\theta)$'s representations into P-Space to capture global data structure information via optimal transport.

\textbf{Global Prototypes Initialization.}
To start with, let us rewrite the deep tabular network $F(\cdot;\theta)$ as $G_h(G_f(\cdot;\theta_f);\theta_h)$, where $G_f(\cdot;\theta_f)$ is the backbone and $G_h(\cdot;\theta_h)$ is the head parameterized by $\theta_f$ and $\theta_h$ respectively. 
To obtain global prototypes initialization, we first train $F(\cdot;\theta)$ by:
\begin{equation}
    \label{equ:org-obj}
    \min\mathcal{L}_{task}(X, Y) = \min_{\theta_f, \theta_h} \mathcal{L}(G_h\left(G_f(X; \theta_f); \theta_h\right), Y),
\end{equation}
where $\mathcal{L}$ is the loss function, and then propose applying K-Means clustering ~\citep{hartigan1979algorithm} to the output of the trained backbone $G_f(X; \theta_{f})$:
\begin{equation}
\label{K-means}
\begin{aligned}
    \min_{C\in \mathbb{R}^{K\times d}} \frac{1}{n}\sum_{i=1}^{n} \min_{\tilde{y_i}\in \{0,1\}^K} \|G_f(x_i; \theta_{f}) - \tilde{y_i}^\mathrm{T} C\|,\  \text{ subject to } \tilde{y_i}^\mathrm{T} \textbf{1}_K = 1, 
\end{aligned}
\end{equation}
where $\textbf{1}_K \in \mathbb{R}^K$ is a vector of ones, $\tilde{y_i}$ is the centroid index and $C$ is the centroid matrix.
The centroid matrix $C$ would be served as the initialization of prototypes $\mathcal{B}$ to guide the stage (ii) training process of \method.
The advantage of using K-Means to generate the initialization of global prototypes lies in (a) enabling the preservation of global data structure information from the trained $F(\cdot;\theta)$ in stage (i), despite the presence of noise,  and (b) facilitating a faster convergence of the stage (ii) training process compared to random initialization.


\textbf{Representation Projecting with Global Data Information via Optimal Transport.} 
After constructing the global prototypes $\mathcal{B}$, we re-initialize the parameters of $F(\cdot;\theta)$ and start to project $F(\cdot;\theta)$'s representation into P-Space via \method to capture global data structure information.
To obtain the projection of $i$-th instance representation $G_f(x_i; \theta_f)$, we use a shared estimator $\phi(\cdot;\gamma)$ with learnable $\gamma$ to calculate its coordinate $r_i$ by $\phi(G_f(x_i; \theta_f);\gamma)$.
Mathematically, our formulation for the $i$-th projection representation distribution takes the following form: $Q_i = \sum_{k=1}^{K} r_{i}^{k} \delta_{\beta_{k}}$.
On the other hand, the $i$-th latent representation $G_f(x_i;\theta_{f})$ could be viewed as the empirical distribution over single sample representation: $P_i = \delta_{G_f(x_i;\theta_{f})}$.

Since all samples are sampled from the same data distribution, it is reasonable to assume that there exists shared structural information among these samples. 
To capture the shared global data structure information, we formulate the representation projecting as the process of extracting instance-wise data information by $G_f(x_i;\theta_{f})$ to $P_i$, and then pushing $P_i$ towards $Q_i$ to encourage each prototype $\beta_k$ to capture the shared \textit{global data structure information}, a process facilitated by leveraging the optimal transport (OT) distance:
\begin{equation}
\label{ot_loss}
\begin{aligned}
    \min\mathcal{L}_{projecting}(X, \mathcal{B}) = \min\frac{1}{n}\sum_{i=1}^{n} \text{OT}(P_i, Q_i) = \min_{\theta_{f}, \gamma, \mathcal{B}}  \frac{1}{n}\sum_{i=1}^{n} \min_{\textbf{T}_i\in \Pi (P_i, Q_i)} \langle \textbf{T}_i, \textbf{C}_i \rangle, \\
\end{aligned}
\end{equation}
where $\textbf{C}$ is the transport cost matrix calculated by cosine distance and $\textbf{T}$ is the transport probability matrix, please refer to Section~\ref{sec: background} for more details about OT. 
We provide the more detailed explanation of the optimization process of Eq.~\ref{ot_loss} in Appendix~\ref{explanation of Eq. 4}.
In contrast to the latent space of $F(\cdot;\theta)$, the P-Space consists of explicitly defined basis vectors denoted as global prototypes, thus it is more accurate to give predictions on representations within P-Space. 
We incorporate the label information by re-defining Eq.~\ref{equ:org-obj}:
\begin{equation}
\label{task specific loss}
\begin{aligned}
    \min\mathcal{L}_{task}(X, Y) = \min_{\theta_f, \theta_h, \gamma, \mathcal{B}} \frac{1}{n}\sum_{i=1}^n \mathcal{L}(G_h(\sum_{k=1}^{K} r_i^k \beta_k;\theta_h), y_i),
\end{aligned}
\end{equation}

By representing each projected representation using the shared basis vectors (global prototypes $\mathcal{B}$) within the P-Space, we can effectively model this \textit{global data structure information} to effectively solve one of the deep tabular network's inherent characteristic: sample localization.
The representation entanglement problem still exists within P-Space, and we will introduce the designed constraints to make the representation disentangled in the following section.

\subsection{Constraints for \method via Coordinates Diversifying and Prototype Matrix Orthogonalization} \label{sec:ptarl_constraints}
\textbf{P-Space Coordinates Diversifying Constraint}. 
Directly optimizing the above process could potentially result in a collapse of coordinates generating, which entails all projection representations exhibiting similar coordinates within P-Space. To alleviate it, we design a constraint to diversifying the coordinates towards global prototypes within P-Space to separate the representation into several disjoint regions, where representations with similar labels are in the same regions.
Specifically, for classification, we categorize samples within the same class as positive pairs ~\citep{khosla2020supervised}, while those from different classes form negative pairs. 
In the context of regression, we segment the labels within a minibatch into \(t\) sub-intervals as indicated by $t = 1 + \log(n_b)$, 
where $n_b$ is bachsize.
Samples residing within the same bin are regarded as positive pairs, while those from distinct bins are regarded as negative pairs, thereby facilitating the formation of pairings. 
This process is achieved by minimizing:
\begin{equation}
\label{diversity_loss}
\begin{aligned}
    \mathcal{L}_{diversifying}(X) = -\sum_{i=1}^{n_b}\sum_{j=1}^{n_b}\textbf{1}\{y_i, y_j \in \text{positive pair}\} \log{\frac{\exp{(\cos(r_i, r_j))}}{\sum_{i=1}^{n_b}\sum_{j=1}^{n_b} \exp{(\cos(r_i, r_j))}}}
\end{aligned}
\end{equation}

This constraint is motivated by contrastive learning (CL) ~\citep{khosla2020supervised}. 
Distinct from conventional contrastive learning methods in tabular domain that directly augment the variety of sample representations ~\citep{wang2022transtab, bahri2021scarf}, it diversifies the coordinates of latent representations within P-Space to calibrate the entangled latent representations.
In this context, the use of prototypes as basis vectors defines a structured coordinate system in the P-Space, thereby facilitating the enhancement of generating disentangled representations among samples, as opposed to directly optimizing their representations.
To improve computational efficiency, practically, we randomly select 50\% of the samples within a minibatch. 
We posit that this approach can mitigate computational complexity while makes it easier to approximate the model to the optimized state.


\textbf{Global Prototype Matrix Orthogonalization Constraint}. 
Since the P-Space is composed of a set of global prototypes, to better represent the P-Space, these global prototypes should serve as the basis vectors, with each prototype maintaining orthogonal independence from the others. 
The presence of interdependence among prototypes would invariably compromise the representation efficacy of these prototypes.
To ensure the independence of prototypes from one another, the condition of orthogonality must be satisfied. 
This mandates the following approach:
\begin{equation}
\label{regulariz_loss}
\begin{aligned}
    \min \mathcal{L}_{orthogonalization}(\mathcal{B}) = \min (\frac{\|M\|_{1}}{\|M\|_{2}^{2}} + \max(0,|K-\|M\|_{1}|)),
\end{aligned}
\end{equation}

where $M \in [0, 1]^{K\times K}$ and $M_{ij} = \|\cos(\beta_i, \beta_j)\|_1$.
The first term $\frac{\|M\|_{1}}{\|M\|_{2}^{2}}$ forces the $M$ to be sparse, i.e., any element $M_{ij} = \|\cos(\beta_i, \beta_j)\|_1$ to be close to 0 ($\beta_i$ and $\beta_j$ is orthogonal) or 1, while the second term motivates $\|M\|_{1} \to K$, i.e., only $K$ elements close to 1.
Since $M_{ii} = \|\cos(\beta_i, \beta_i)\|_1 = 1, \forall i \in \{1,2,...,K\}$, it would force $M_{ij} = \|\cos(\beta_i, \beta_j)\|_1, \forall i,j \in \{1,2,...,K\}, i \neq j$ to be 0, i.e. each prototype maintains orthogonal independence from the others.

\section{Experiment \& Analysis}
\label{sec: experiment}

\subsection{Experiment Setup}
\textbf{Datasets.}
As described before, it is challenging for deep networks to achieve satisfactory performance due to heterogeneity, complexity, and diversity of tabular data. 
In this paper, we consider a variety of supervised tabular deep learning tasks with heterogeneous features, including binary classification, multiclass classification, and regression.
Specifically, the tabular datasets include: Adult (AD)~\citep{kohavi1996scaling}, Higgs (HI)~\citep{vanschoren2014openml}, Helena (HE)~\citep{guyon2019analysis}, Jannis (JA)~\citep{guyon2019analysis}, ALOI (AL)~\citep{geusebroek2005amsterdam}, California\_housing (CA)~\citep{pace1997sparse}.
The dataset properties are summarized in Appendix~\ref{appendix:datasets details}.
We split each dataset into training, validation and test set by the ratio of
6:2:2. 
For the data pre-processing, please refer to Appendix~\ref{appendix:datasets details} for more details.

\textbf{Baseline Deep Tabular Models.}
As the \method is a model-agnostic pipeline, we include 6 mainstream deep tabular models to test \method's applicability and effectiveness to different predictors with diverse architectures, which are as follows: MLP~\citep{taud2018multilayer}, ResNet~\citep{he2016deep}, SNN~\citep{klambauer2017self}, DCNV2~\citep{wang2021dcn}, AutoInt~\citep{song2019autoint}, and FT-Transformer~\citep{gorishniy2021revisiting}.
More details could be found in Appendix~\ref{appendix:baseline details}. 

\textbf{\method Details.} 
The \method is a two-stage model-agnostic pipeline that aims to enhance the performance of any deep tabular model $F(\cdot;\theta)$ without altering its internal architecture. 
The first stage is to construct the core of \method, i.e. P-Space, that consists of a set of global prototypes $\mathcal{B}$.
The number of global prototypes $K$ is data-specific, and we set $K$ to the ceil of $\log(N)$, where $N$ is the total number of features. 
The estimator $\phi(\cdot;\gamma)$, which is used to calculate the coordinates of representations within P-Space, is a simple 3-layer fully-connected MLP.
To ensure fairness, in the second stage of training, we inherit the hyperparameters of $F(\cdot;\theta)$ (the learnable $\theta$ would be re-initialized).
We provide the \method workflow in Appendix~\ref{appendix:implementation details}.
Following the common practice of previous studies, we use Root-Mean-Square Error (RMSE) (lower is better) to evaluate the regression tasks, Accuracy (higher is better) to evaluate binary and multiclass classification tasks. 
To reduce the effect of randomness, the reported performance is averaged over 10 independent runs.

\subsection{Empirical Results}
\textbf{\method generally improves deep tabular models' performance.}
From Table~\ref{table:main results} we can observe that \method achieves consistent improvements over the baseline deep models in all settings. 
It achieves a more than 4\% performance improvement for all settings, whether using Accuracy or RMSE as the evaluation metric.
In addition, we conduct Wilcoxon signed-rank test (with $\alpha=0.05$) ~\citep{woolson2007wilcoxon} to measure the improvement significance.
In all settings, the improvement of \method over deep models is statistically significant at the 95\% confidence level.
This demonstrates the superior adaptability and generality of \method to different models and tasks.
In addition, the results also indicate that there is no deep model that consistently outperforms others on all tasks, i.e., the universal winner solution does not exist, which is aligned with the findings in previous works~\citep{gorishniy2021revisiting}. 

\textbf{Ablation results.}
We further conduct ablation study to demonstrate the effectiveness of key components of \method.
Specifically, we conduct a comparison between the deep model coupled with \method and three variants: (i) \method w/o O, that removes the global prototype matrix orthogonalization constraint, (ii) \method w/o O, D, that further removes the P-Space coordinates diversifying constraint and (iii) w/o \method, that is identical to directly training deep models by Eq.~\ref{equ:org-obj}. 
The results in Table~\ref{table:ablation} show that the removal of any of the components degrades the performance of \method.
The comparison between \method w/o O, D and w/o \method indicates that by projecting deep tabular model's representation into P-Space, the shared \textit{global data structure information} are captured by global prototypes to solve the \textit{representation localization} to enhance the representation quality. 
In addition, diversifying the representation coordinates in P-Space and orthogonalizing the global prototypes of P-Space could both enable the generation of \textit{disentangled representations} to alleviate the \textit{representation entanglement} problem.
Besides, we also provide additional ablation tests, including validating the effectiveness of K-Means as the global prototypes initialization method and the effectiveness of Optimal Transport (OT) as the distribution measurement in Table~\ref{table:ablation_ini_meas}.
The result of initialization method is aligned to our explanation for the advantage of K-Means in Section~\ref{sec:ptarl_main}. 
In addition, the result of distribution measurement indicates that compared to Manhattan distance and Euclidean distance, OT measures the minimum distance between two distributions through point by point calculation, which could better capture the data structure between two distributions.
Due to the space limit, we leave more details and results about the above ablation study to Appendix~\ref{appendix: experimental results}.

\begin{table}
\vspace{-1em}
\centering
\captionsetup{font=small}
\caption{Tabular prediction performance of \method over different deep tabular models for different tasks. ``$\uparrow$'' represents higher evaluation metric is better for classification, ``$\downarrow$'' represents lower evaluation metric is better for regression. The best results are highlighted in bold. ``Win'' represents the number of datasets that one scheme achieves the best.}
\label{table:main results}
\scriptsize
\setlength{\tabcolsep}{0.8mm}{
\begin{tabular}{c|cc|cc|cc|cc|cc|cc}
\toprule
               & MLP   & \textbf{+PTaRL}  & DCNV2  & \textbf{+PTaRL}   & SNN   & \textbf{+PTaRL}  & ResNet   & \textbf{+PTaRL}  & AutoInt   & \textbf{+PTaRL} & FT-Trans   & \textbf{+PTaRL}  \\ \midrule
AD $\uparrow$  & 0.825 & \textbf{0.868}   & 0.826  & \textbf{0.867}    & 0.825 & \textbf{0.859}   & 0.813    & \textbf{0.862}   & 0.823     & \textbf{0.871}  & 0.827      & \textbf{0.871}   \\
HI $\uparrow$  & 0.681 & \textbf{0.723}   & 0.681  & \textbf{0.731}    & 0.69  & \textbf{0.724}   & 0.682    & \textbf{0.729}   & 0.685     & \textbf{0.738}  & 0.687      & \textbf{0.738}   \\
HE $\uparrow$  & 0.352 & \textbf{0.396}   & 0.34   & \textbf{0.389}    & 0.338 & \textbf{0.394}   & 0.354    & \textbf{0.399}   & 0.338     & \textbf{0.396}  & 0.352      & \textbf{0.397}   \\
JA $\uparrow$  & 0.672 & \textbf{0.71}    & 0.662  & \textbf{0.723}    & 0.689 & \textbf{0.732}   & 0.666    & \textbf{0.723}   & 0.653     & \textbf{0.722}  & 0.689      & \textbf{0.738}   \\
AL $\uparrow$  & 0.917 & \textbf{0.964}   & 0.905  & \textbf{0.959}    & 0.917 & \textbf{0.961}   & 0.919    & \textbf{0.964}   & 0.894     & \textbf{0.955}  & 0.924      & \textbf{0.97}    \\
CA $\downarrow$& 0.518 & \textbf{0.489}   & 0.502  & \textbf{0.465}    & 0.898 & \textbf{0.631}   & 0.537    & \textbf{0.498}   & 0.507     & \textbf{0.464}  & 0.486      & \textbf{0.448}   \\ \hline
Win            & 0     & \textbf{6}       & 0      & \textbf{6}        & 0     & \textbf{6}       & 0        & \textbf{6}       & 0         & \textbf{6}      & 0          & \textbf{6}       \\        
\bottomrule
\end{tabular}}
\end{table}

\begin{table}
\vspace{-1em}
\centering
\captionsetup{font=small}
\caption{Ablation results on the effects of different components in \method. The experiment is conducted on FT-Transformer. The best results are highlighted in bold.}
\scriptsize
\label{table:ablation}
\setlength{\tabcolsep}{2.8mm}{
\begin{tabular}{c|cccccc|c}
\toprule
                                                 & AD $\uparrow$  & HI $\uparrow$  & HE $\uparrow$  & JA $\uparrow$  & AL $\uparrow$  & CA $\downarrow$     & Win         \\ \midrule
FT-Transformer + \method                         & \textbf{0.871} & \textbf{0.738} & \textbf{0.397} & \textbf{0.738} & \textbf{0.97}  & \textbf{0.448}      & 6           \\
\method w/o O                                            & 0.859          & 0.722          & 0.383          & 0.725          & 0.96           & 0.453               & 0           \\
\method w/o O, D                                         & 0.841          & 0.704          & 0.369          & 0.702          & 0.94           & 0.466               & 0           \\
w/o \method                                      & 0.827          & 0.687          & 0.352          & 0.689          & 0.924          & 0.486               & 0           \\
\bottomrule
\end{tabular}}
\vspace{-1em}
\end{table}

\begin{table}
\vspace{-1em}
\centering
\captionsetup{font=small}
\caption{Ablation results on the effects of different prototype initialization and distribution measurement in \method. The experiment is conducted on FT-Transformer. The best results are highlighted in bold.}
\scriptsize
\label{table:ablation_ini_meas}
\setlength{\tabcolsep}{2.8mm}{
\begin{tabular}{c|cccccc}
\toprule
Initialization             & AD $\uparrow$  & HI $\uparrow$   & HE $\uparrow$   & JA $\uparrow$   & AL $\uparrow$   & CA $\downarrow$   \\ \hline
\method w/ K-Means (Ours)  & \textbf{0.871} & \textbf{0.738}  & \textbf{0.397}  & \textbf{0.738}  & \textbf{0.97}   & \textbf{0.448}    \\
\method w/ Random          & 0.863          & 0.721           & 0.389           & 0.729           & 0.951           & 0.462             \\ 
\bottomrule
\toprule
Distribution measurement              & AD $\uparrow$   & HI $\uparrow$    & HE $\uparrow$   & JA $\uparrow$   & AL $\uparrow$    & CA $\downarrow$ \\ \hline
\method w/ OT (Ours)                  & \textbf{0.871}  & \textbf{0.738}   & \textbf{0.397}  & \textbf{0.738}  & \textbf{0.97}    & \textbf{0.448}  \\
\method w/ Manhattan distance         & 0.859           & 0.729            & 0.382           & 0.721           & 0.953            & 0.454           \\  
\method w/ Euclidean distance         & 0.862           & 0.713            & 0.390           & 0.724           & 0.949            & 0.459           \\ 
\bottomrule
\end{tabular}}
\vspace{-1em}
\end{table}

\textbf{Computational efficiency and sensitivity analysis.} 
In our paper, while Optimal Transport (OT) demonstrates strong computational capabilities for measuring the minimum distance between two distributions, it does not significantly increase computational complexity. We include the computational efficiency details in Appendix~\ref{Appendix: computational efficiency details}.
In addition, we also incorporate the sensitivity analysis for the weights of different loss functions in Appendix~\ref{Appendix: sensitivity analysis}.
Since the core of \method is the constructed P-Space, we further explore \method's performance under different global prototypes number in Appendix~\ref{Appendix: sensitivity analysis}.

\begin{figure}[h]
\centering
\includegraphics[width=0.8\linewidth]{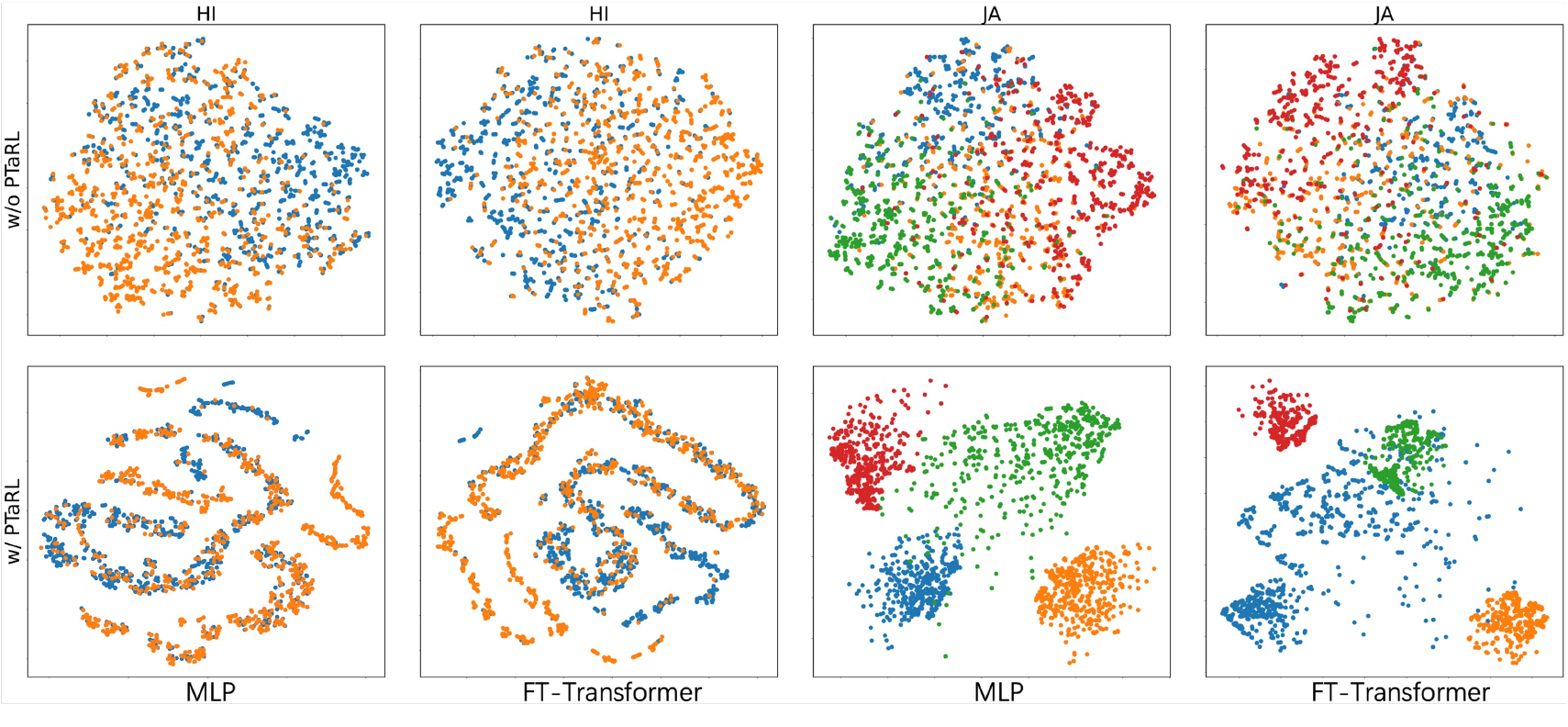}
\captionsetup{font=small}
\caption{Visualization of learned representations of deep tabular models w/ and w/o \method.
}
\label{fig:PTaRL_visualization}
\vspace{-2em}
\end{figure}
\subsection{\method procedure analysis and visualization}
\vspace{-1em}
\textbf{\method enables the generation of disentangled representation.} 
Fig.~\ref{fig:PTaRL_visualization} shows the learned representation of deep model $F(\cdot;\theta)$ w/ and w/o \method on binary and multiclass classification tasks.
The first row shows that different deep models suffer from the \textit{representation entanglement} problem and this is aligned with our motivation.
With \method, representations within P-Space are separated into several disjoint regions, where representations with similar labels are in the same regions. 
This demonstrates \method' ability to generate \textit{disentangled representations} for tabular deep learning.
\begin{minipage}{0.47\textwidth} 
\textbf{\method enables P-Space coordinates diversifying.} To validate whether the learned coordinates of representation within P-Space have been diversified, we average the coordinates of data points which are belong to the same category and visualize them in Fig.~\ref{fig:coordinates_diversifying}. 
Without the Diversifying Constraint, coordinates of different categories appear similar, but incorporating the constraint enhances the diversity between category coordinates.
\textbf{\method enables the orthogonalization of global prototypes within P-Space.}
To explore the structure of constructed P-Space, we visualize the relation between any $(\beta_i, \beta_j)$ by calculating $\|\cos(\beta_i, \beta_j)\|_1$ in Fig.~\ref{fig:prototype_orthogonalization}. Each prototype maintains orthogonal independence from the 
others, which demonstrates the effectiveness of prototype matrix orthogonalization constraint. 
\end{minipage}
\hspace{0.1cm}
\begin{minipage}{0.5\textwidth} 
    \includegraphics[width=\linewidth]{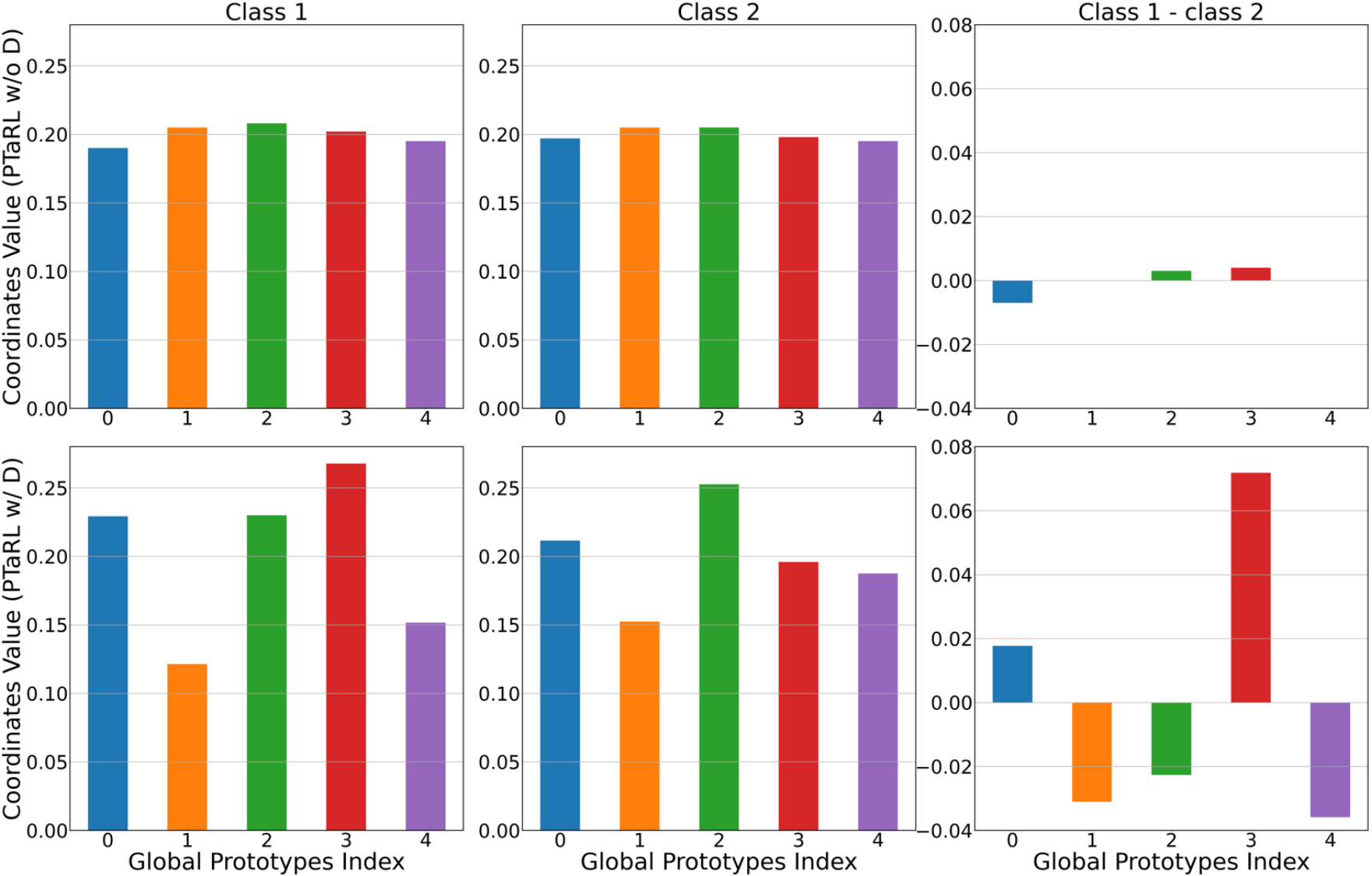}
    \captionsetup{font=small}
    \captionsetup{skip=0pt}
    \captionof{figure}{\label{fig:coordinates_diversifying}P-Space coordinates diversifying visualization of FT-Transformer on HI w/o and w/ Coordinates Diversifying Constraint (D). The first column and second column correspond to the average coordinates values of two different categories, the third column represents the difference of the two categories.}
\end{minipage}%

\begin{figure}[h]
\centering
\includegraphics[width=0.9\linewidth]{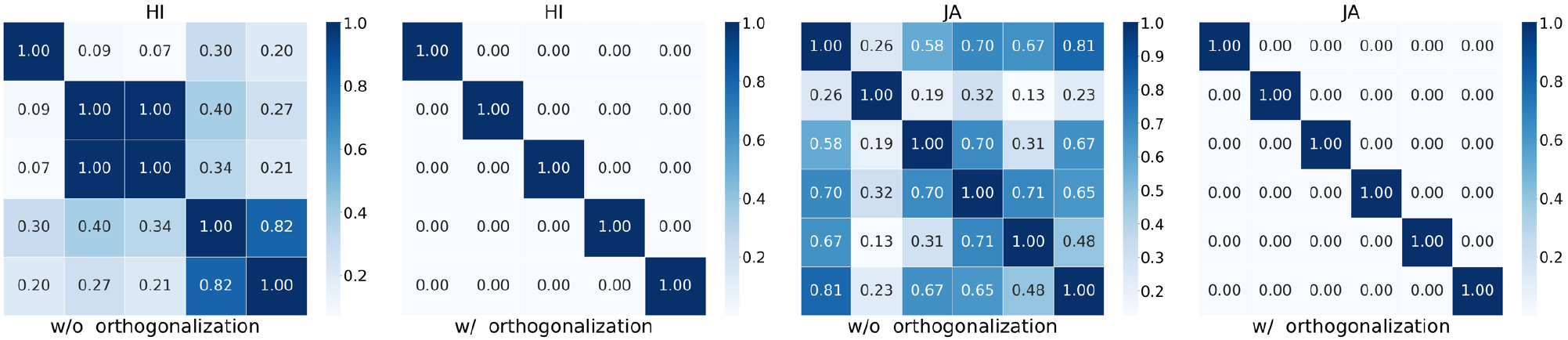}
\captionsetup{font=small}
\caption{Global prototypes orthogonalization visualization of MLP on two different tasks. 
}
\label{fig:prototype_orthogonalization}
\end{figure}


\section{Conclusion}
\vspace{-1em}
In this paper, we have investigated the learned patterns of deep tabular models and identified two inherent representation challenges hindering satisfactory performance, i.e. \textit{sample localization} and \textit{representation entanglement}.
To handle these challenges, we proposed \method, a prototype-based tabular representation learning pipeline that can be coupled with any deep tabular model to enhance the representation quality. 
The core of the \method is the constructed P-Space, that consists of a set of global prototypes.
\method mainly involves two stages, i.e. global prototype generation and projecting representations into P-Space, to capture the \textit{global data structure information}.
Besides, two constraints are designed to \textit{disentangle} the projected representations within P-Space.
The empirical results on various real world tasks demonstrated the effectiveness of \method for tabular deep learning.
Our work can shed some light on developing better algorithms for similar tasks.







\subsubsection*{Acknowledgments}
This work is supported by the National Natural Science Foundation of China (No.61976102, No.U19A2065, No.62306125) and Key R\&D Program of the Ministry of Science
and Technology, China (2023YFF0905400).

\bibliography{iclr2024_conference}

\begin{thebibliography}{45}
\providecommand{\natexlab}[1]{#1}
\providecommand{\url}[1]{\texttt{#1}}
\expandafter\ifx\csname urlstyle\endcsname\relax
  \providecommand{\doi}[1]{doi: #1}\else
  \providecommand{\doi}{doi: \begingroup \urlstyle{rm}\Url}\fi

\bibitem[Altschuler et~al.(2017)Altschuler, Niles-Weed, and Rigollet]{altschuler2017near}
Jason Altschuler, Jonathan Niles-Weed, and Philippe Rigollet.
\newblock Near-linear time approximation algorithms for optimal transport via sinkhorn iteration.
\newblock \emph{Advances in neural information processing systems}, 30, 2017.

\bibitem[Assefa et~al.(2020)Assefa, Dervovic, Mahfouz, Tillman, Reddy, and Veloso]{assefa2020generating}
Samuel~A Assefa, Danial Dervovic, Mahmoud Mahfouz, Robert~E Tillman, Prashant Reddy, and Manuela Veloso.
\newblock Generating synthetic data in finance: opportunities, challenges and pitfalls.
\newblock \emph{Proceedings of the First ACM International Conference on AI in Finance}, pp.\  1--8, 2020.

\bibitem[Bahri et~al.(2021)Bahri, Jiang, Tay, and Metzler]{bahri2021scarf}
Dara Bahri, Heinrich Jiang, Yi~Tay, and Donald Metzler.
\newblock Scarf: Self-supervised contrastive learning using random feature corruption.
\newblock \emph{arXiv preprint arXiv:2106.15147}, 2021.

\bibitem[Chen et~al.(2022)Chen, Liao, Wan, Chen, and Wu]{chen2022danets}
Jintai Chen, Kuanlun Liao, Yao Wan, Danny~Z Chen, and Jian Wu.
\newblock Danets: Deep abstract networks for tabular data classification and regression.
\newblock In \emph{Proceedings of the AAAI Conference on Artificial Intelligence}, 2022.

\bibitem[Chen \& Guestrin(2016)Chen and Guestrin]{chen2016xgboost}
Tianqi Chen and Carlos Guestrin.
\newblock Xgboost: A scalable tree boosting system.
\newblock \emph{Proceedings of the 22nd acm sigkdd international conference on knowledge discovery and data mining}, pp.\  785--794, 2016.

\bibitem[Chizat et~al.(2020)Chizat, Roussillon, L{\'e}ger, Vialard, and Peyr{\'e}]{chizat2020faster}
Lenaic Chizat, Pierre Roussillon, Flavien L{\'e}ger, Fran{\c{c}}ois-Xavier Vialard, and Gabriel Peyr{\'e}.
\newblock Faster wasserstein distance estimation with the sinkhorn divergence.
\newblock \emph{Advances in Neural Information Processing Systems}, 33:\penalty0 2257--2269, 2020.

\bibitem[Cuturi(2013)]{cuturi2013sinkhorn}
Marco Cuturi.
\newblock Sinkhorn distances: Lightspeed computation of optimal transport.
\newblock \emph{Advances in neural information processing systems}, 26, 2013.

\bibitem[Devlin et~al.(2018)Devlin, Chang, Lee, and Toutanova]{devlin2018bert}
Jacob Devlin, Ming-Wei Chang, Kenton Lee, and Kristina Toutanova.
\newblock Bert: Pre-training of deep bidirectional transformers for language understanding.
\newblock \emph{arXiv preprint arXiv:1810.04805}, 2018.

\bibitem[Du et~al.(2021)Du, Gao, Chen, Jia, Wang, Zhang, Han, and Zhang]{du2021tabularnet}
Lun Du, Fei Gao, Xu~Chen, Ran Jia, Junshan Wang, Jiang Zhang, Shi Han, and Dongmei Zhang.
\newblock Tabularnet: A neural network architecture for understanding semantic structures of tabular data.
\newblock \emph{Proceedings of the 27th ACM SIGKDD Conference on Knowledge Discovery \& Data Mining}, pp.\  322--331, 2021.

\bibitem[Frosch et~al.(2010)Frosch, Grande, Tarn, and Kravitz]{frosch2010decade}
Dominick~L Frosch, David Grande, Derjung~M Tarn, and Richard~L Kravitz.
\newblock A decade of controversy: balancing policy with evidence in the regulation of prescription drug advertising.
\newblock \emph{American Journal of Public Health}, 100\penalty0 (1):\penalty0 24--32, 2010.

\bibitem[Geusebroek et~al.(2005)Geusebroek, Burghouts, and Smeulders]{geusebroek2005amsterdam}
Jan-Mark Geusebroek, Gertjan~J Burghouts, and Arnold~WM Smeulders.
\newblock The amsterdam library of object images.
\newblock \emph{International Journal of Computer Vision}, 61:\penalty0 103--112, 2005.

\bibitem[Gorishniy et~al.(2021)Gorishniy, Rubachev, Khrulkov, and Babenko]{gorishniy2021revisiting}
Yury Gorishniy, Ivan Rubachev, Valentin Khrulkov, and Artem Babenko.
\newblock Revisiting deep learning models for tabular data.
\newblock \emph{Advances in Neural Information Processing Systems}, 34:\penalty0 18932--18943, 2021.

\bibitem[Guyon et~al.(2019)Guyon, Sun-Hosoya, Boull{\'e}, Escalante, Escalera, Liu, Jajetic, Ray, Saeed, Sebag, et~al.]{guyon2019analysis}
Isabelle Guyon, Lisheng Sun-Hosoya, Marc Boull{\'e}, Hugo~Jair Escalante, Sergio Escalera, Zhengying Liu, Damir Jajetic, Bisakha Ray, Mehreen Saeed, Mich{\`e}le Sebag, et~al.
\newblock Analysis of the automl challenge series.
\newblock \emph{Automated Machine Learning}, 177, 2019.

\bibitem[Hartigan \& Wong(1979)Hartigan and Wong]{hartigan1979algorithm}
John~A Hartigan and Manchek~A Wong.
\newblock Algorithm as 136: A k-means clustering algorithm.
\newblock \emph{Journal of the royal statistical society. series c (applied statistics)}, 28\penalty0 (1):\penalty0 100--108, 1979.

\bibitem[He et~al.(2016)He, Zhang, Ren, and Sun]{he2016deep}
Kaiming He, Xiangyu Zhang, Shaoqing Ren, and Jian Sun.
\newblock Deep residual learning for image recognition.
\newblock \emph{Proceedings of the IEEE conference on computer vision and pattern recognition}, pp.\  770--778, 2016.

\bibitem[Hernandez et~al.(2022)Hernandez, Epelde, Alberdi, Cilla, and Rankin]{hernandez2022synthetic}
Mikel Hernandez, Gorka Epelde, Ane Alberdi, Rodrigo Cilla, and Debbie Rankin.
\newblock Synthetic data generation for tabular health records: A systematic review.
\newblock \emph{Neurocomputing}, 493:\penalty0 28--45, 2022.

\bibitem[Hoang et~al.(2020)Hoang, Lam, Low, and Jaillet]{hoang20b}
Nghia Hoang, Thanh Lam, Bryan Kian~Hsiang Low, and Patrick Jaillet.
\newblock Learning task-agnostic embedding of multiple black-box experts for multi-task model fusion.
\newblock \emph{International Conference on Machine Learning}, 119:\penalty0 4282--4292, 2020.

\bibitem[Huang et~al.(2012)Huang, Socher, Manning, and Ng]{huang2012improving}
Eric~H Huang, Richard Socher, Christopher~D Manning, and Andrew~Y Ng.
\newblock Improving word representations via global context and multiple word prototypes.
\newblock \emph{Proceedings of the 50th annual meeting of the association for computational linguistics (Volume 1: Long papers)}, pp.\  873--882, 2012.

\bibitem[Huang et~al.(2020)Huang, Khetan, Cvitkovic, and Karnin]{huang2020tabtransformer}
Xin Huang, Ashish Khetan, Milan Cvitkovic, and Zohar Karnin.
\newblock Tabtransformer: Tabular data modeling using contextual embeddings.
\newblock \emph{arXiv preprint arXiv:2012.06678}, 2020.

\bibitem[Ke et~al.(2017)Ke, Meng, Finley, Wang, Chen, Ma, Ye, and Liu]{ke2017lightgbm}
Guolin Ke, Qi~Meng, Thomas Finley, Taifeng Wang, Wei Chen, Weidong Ma, Qiwei Ye, and Tie-Yan Liu.
\newblock Lightgbm: A highly efficient gradient boosting decision tree.
\newblock \emph{Advances in neural information processing systems}, 30, 2017.

\bibitem[Khosla et~al.(2020)Khosla, Teterwak, Wang, Sarna, Tian, Isola, Maschinot, Liu, and Krishnan]{khosla2020supervised}
Prannay Khosla, Piotr Teterwak, Chen Wang, Aaron Sarna, Yonglong Tian, Phillip Isola, Aaron Maschinot, Ce~Liu, and Dilip Krishnan.
\newblock Supervised contrastive learning.
\newblock \emph{Advances in neural information processing systems}, 33:\penalty0 18661--18673, 2020.

\bibitem[Klambauer et~al.(2017)Klambauer, Unterthiner, Mayr, and Hochreiter]{klambauer2017self}
G{\"u}nter Klambauer, Thomas Unterthiner, Andreas Mayr, and Sepp Hochreiter.
\newblock Self-normalizing neural networks.
\newblock \emph{Advances in neural information processing systems}, 30, 2017.

\bibitem[Kohavi et~al.(1996)]{kohavi1996scaling}
Ron Kohavi et~al.
\newblock Scaling up the accuracy of naive-bayes classifiers: A decision-tree hybrid.
\newblock \emph{KDD}, 96:\penalty0 202--207, 1996.

\bibitem[Li et~al.(2021)Li, Jampani, Sevilla-Lara, Sun, Kim, and Kim]{li2021adaptive}
Gen Li, Varun Jampani, Laura Sevilla-Lara, Deqing Sun, Jonghyun Kim, and Joongkyu Kim.
\newblock Adaptive prototype learning and allocation for few-shot segmentation.
\newblock \emph{Proceedings of the IEEE/CVF conference on computer vision and pattern recognition}, pp.\  8334--8343, 2021.

\bibitem[Nauta et~al.(2021)Nauta, Van~Bree, and Seifert]{nauta2021neural}
Meike Nauta, Ron Van~Bree, and Christin Seifert.
\newblock Neural prototype trees for interpretable fine-grained image recognition.
\newblock \emph{Proceedings of the IEEE/CVF Conference on Computer Vision and Pattern Recognition}, pp.\  14933--14943, 2021.

\bibitem[Pace \& Barry(1997)Pace and Barry]{pace1997sparse}
R~Kelley Pace and Ronald Barry.
\newblock Sparse spatial autoregressions.
\newblock \emph{Statistics \& Probability Letters}, 33\penalty0 (3):\penalty0 291--297, 1997.

\bibitem[Pedregosa et~al.(2011)Pedregosa, Varoquaux, Gramfort, Michel, Thirion, Grisel, Blondel, Prettenhofer, Weiss, Dubourg, et~al.]{pedregosa2011scikit}
Fabian Pedregosa, Ga{\"e}l Varoquaux, Alexandre Gramfort, Vincent Michel, Bertrand Thirion, Olivier Grisel, Mathieu Blondel, Peter Prettenhofer, Ron Weiss, Vincent Dubourg, et~al.
\newblock Scikit-learn: Machine learning in python.
\newblock \emph{the Journal of machine Learning research}, 12:\penalty0 2825--2830, 2011.

\bibitem[Peyr{\'e} et~al.(2017)Peyr{\'e}, Cuturi, et~al.]{peyre2017computational}
Gabriel Peyr{\'e}, Marco Cuturi, et~al.
\newblock Computational optimal transport.
\newblock \emph{Center for Research in Economics and Statistics Working Papers}, pp.\  2017--86, 2017.

\bibitem[Shwartz-Ziv \& Armon(2022)Shwartz-Ziv and Armon]{shwartz2022tabular}
Ravid Shwartz-Ziv and Amitai Armon.
\newblock Tabular data: Deep learning is not all you need.
\newblock \emph{Information Fusion}, 81:\penalty0 84--90, 2022.

\bibitem[Song et~al.(2019)Song, Shi, Xiao, Duan, Xu, Zhang, and Tang]{song2019autoint}
Weiping Song, Chence Shi, Zhiping Xiao, Zhijian Duan, Yewen Xu, Ming Zhang, and Jian Tang.
\newblock Autoint: Automatic feature interaction learning via self-attentive neural networks.
\newblock \emph{Proceedings of the 28th ACM international conference on information and knowledge management}, pp.\  1161--1170, 2019.

\bibitem[Su et~al.(2012)Su, Yan, and Tsai]{su2012linear}
Xiaogang Su, Xin Yan, and Chih-Ling Tsai.
\newblock Linear regression.
\newblock \emph{Wiley Interdisciplinary Reviews: Computational Statistics}, 4\penalty0 (3):\penalty0 275--294, 2012.

\bibitem[Taud \& Mas(2018)Taud and Mas]{taud2018multilayer}
Hind Taud and JF~Mas.
\newblock Multilayer perceptron (mlp).
\newblock \emph{Geomatic approaches for modeling land change scenarios}, pp.\  451--455, 2018.

\bibitem[Van~der Maaten \& Hinton(2008)Van~der Maaten and Hinton]{van2008visualizing}
Laurens Van~der Maaten and Geoffrey Hinton.
\newblock Visualizing data using t-sne.
\newblock \emph{Journal of machine learning research}, 9\penalty0 (11), 2008.

\bibitem[Vanschoren et~al.(2014)Vanschoren, Van~Rijn, Bischl, and Torgo]{vanschoren2014openml}
Joaquin Vanschoren, Jan~N Van~Rijn, Bernd Bischl, and Luis Torgo.
\newblock Openml: networked science in machine learning.
\newblock \emph{ACM SIGKDD Explorations Newsletter}, 15\penalty0 (2):\penalty0 49--60, 2014.

\bibitem[Wang et~al.(2021)Wang, Shivanna, Cheng, Jain, Lin, Hong, and Chi]{wang2021dcn}
Ruoxi Wang, Rakesh Shivanna, Derek Cheng, Sagar Jain, Dong Lin, Lichan Hong, and Ed~Chi.
\newblock Dcn v2: Improved deep \& cross network and practical lessons for web-scale learning to rank systems.
\newblock \emph{Proceedings of the web conference 2021}, pp.\  1785--1797, 2021.

\bibitem[Wang \& Sun(2022)Wang and Sun]{wang2022transtab}
Zifeng Wang and Jimeng Sun.
\newblock Transtab: Learning transferable tabular transformers across tables.
\newblock \emph{Advances in Neural Information Processing Systems}, 35:\penalty0 2902--2915, 2022.

\bibitem[Woolson(2007)]{woolson2007wilcoxon}
Robert~F Woolson.
\newblock Wilcoxon signed-rank test.
\newblock \emph{Wiley encyclopedia of clinical trials}, pp.\  1--3, 2007.

\bibitem[Wright(1995)]{wright1995logistic}
Raymond~E Wright.
\newblock \emph{Logistic regression.}
\newblock American Psychological Association, 1995.

\bibitem[Xu et~al.(2024)Xu, Huang, Hu, Li, Gilani, Goan, and Liu]{xu2024bishop}
Chenwei Xu, Yu-Chao Huang, Jerry Yao-Chieh Hu, Weijian Li, Ammar Gilani, Hsi-Sheng Goan, and Han Liu.
\newblock Bishop: Bi-directional cellular learning for tabular data with generalized sparse modern hopfield model.
\newblock \emph{arXiv preprint arXiv:2404.03830}, 2024.

\bibitem[Yang et~al.(2018)Yang, Zhang, Yin, and Liu]{yang2018robust}
Hong-Ming Yang, Xu-Yao Zhang, Fei Yin, and Cheng-Lin Liu.
\newblock Robust classification with convolutional prototype learning.
\newblock \emph{Proceedings of the IEEE conference on computer vision and pattern recognition}, pp.\  3474--3482, 2018.

\bibitem[Ye et~al.(2023)Ye, Liu, Shen, Cao, Zheng, Gui, Zhang, Chang, and Bian]{ye2023uadb}
Hangting Ye, Zhining Liu, Xinyi Shen, Wei Cao, Shun Zheng, Xiaofan Gui, Huishuai Zhang, Yi~Chang, and Jiang Bian.
\newblock Uadb: Unsupervised anomaly detection booster.
\newblock \emph{arXiv preprint arXiv:2306.01997}, 2023.

\bibitem[Zalmout \& Li(2022)Zalmout and Li]{zalmout2022prototype}
Nasser Zalmout and Xian Li.
\newblock Prototype-representations for training data filtering in weakly-supervised information extraction.
\newblock pp.\  467--474, 2022.

\bibitem[Zhou et~al.(2020)Zhou, Tao, Pengxin, Shi, and Dongmei]{zhou2020table2analysis}
Mengyu Zhou, Wang Tao, Ji~Pengxin, Han Shi, and Zhang Dongmei.
\newblock Table2analysis: Modeling and recommendation of common analysis patterns for multi-dimensional data.
\newblock \emph{Proceedings of the AAAI Conference on Artificial Intelligence}, 34\penalty0 (01):\penalty0 320--328, 2020.

\bibitem[Zhou et~al.(2022)Zhou, Wang, Konukoglu, and Van~Gool]{zhou2022rethinking}
Tianfei Zhou, Wenguan Wang, Ender Konukoglu, and Luc Van~Gool.
\newblock Rethinking semantic segmentation: A prototype view.
\newblock \emph{Proceedings of the IEEE/CVF Conference on Computer Vision and Pattern Recognition}, pp.\  2582--2593, 2022.

\bibitem[Zhu et~al.(2023)Zhu, Shi, Erickson, Li, Karypis, and Shoaran]{zhu2023xtab}
Bingzhao Zhu, Xingjian Shi, Nick Erickson, Mu~Li, George Karypis, and Mahsa Shoaran.
\newblock Xtab: Cross-table pretraining for tabular transformers.
\newblock \emph{arXiv preprint arXiv:2305.06090}, 2023.

\end{thebibliography}
\bibliographystyle{iclr2024_conference}

\appendix
\newpage
\section{Appendix}
In this section, we provide details of the datasets, baseline methods, the implementation of our method, comprehensive experimental results and visualizations. More detailed information is available at \url{https://github.com/HangtingYe/PTaRL}.

\subsection{Datasets details.}
\label{appendix:datasets details}

\textbf{Data pre-processing.}
Due to the property of neural networks, data pre-processing is important, especially for tabular data. 
To handle categorical features, we adopt an integer encoding scheme, where each category within a column is uniquely mapped to an integer to index the embedding in lookup table. 
Furthermore, we maintain consistent embedding dimensions for all categorical features.
For numerical features, we apply column-wise normalization method.
In regression tasks, we also apply the normalization to the labels. 
To ensure fair comparisons, we adhere to identical preprocessing procedures across all deep networks for each dataset.
Following ~\citep{gorishniy2021revisiting}, we use the quantile transformation from the Scikit-learn library ~\citep{pedregosa2011scikit}. We apply Standardization to HE and AL. The latter one represents image data, and standardization is a common practice in computer vision. 

\begin{table}[h]
    \centering
	\caption{Tabular data properties. Accuracy is used for binary and multiclass classification, RMSE denotes Root Mean Square Error for regression.}
    \setlength{\tabcolsep}{1.8mm}{
    \begin{tabular}{ccccccccccc}
        \toprule
                                   & AD         & HI         & HE        & JA        & AL       & CA    \\
        \midrule
        Objects                    & 48842      & 98050      & 65196     & 83733     & 108000   & 20640 \\
        Numerical, Categorical     & 6, 8       & 28, 0      & 27, 0     & 54, 0     & 128, 0   & 8, 0  \\
        Classes                    & 2          & 2          & 100       & 4         & 1000     & -     \\
        metric                     & Accuracy   & Accuracy   & Accuracy  & Accuracy  & Accuracy & RMSE  \\
        \bottomrule
    \end{tabular}}
\end{table}

\subsection{Baseline deep tabular models details.}
\label{appendix:baseline details}
\begin{itemize}[leftmargin=*]
\item 
MLP~\citep{taud2018multilayer}. 
\item 
DCNV2 ~\citep{wang2021dcn}. Consists of an MLP-like module and the feature crossing
module (a combination of linear layers and multiplications).
\item 
SNN ~\citep{klambauer2017self}. An MLP-like architecture with the SELU activation that
enables training deeper models.
\item 
ResNet ~\citep{he2016deep}. The key innovation is the use of residual connections, also known as skip connections or shortcut connections. These connections enable the network to effectively train very deep neural networks, which was challenging before due to the vanishing gradient problem. 
In this paper, we use the ResNet version introduced by ~\citep{gorishniy2021revisiting}.
\item 
AutoInt ~\citep{song2019autoint}. Transforms features to embeddings and applies a series of
attention-based transformations to the embeddings. 
\item 
FT-Transformer~\citep{gorishniy2021revisiting}. FT-Transformer is introduced by ~\citep{gorishniy2021revisiting} to further improved AutoInt through better token embeddings.
\end{itemize}

\subsection{\method workflow.}
\label{appendix:implementation details}

\begin{algorithm}
\caption{\textsc{PTaRL} algorithm workflow.}
\label{algorithm}
\begin{algorithmic}[1]
\INPUT Input data $D=\{X, Y\}$, deep tabular model $F(\cdot;\theta) = G_h\left(G_f(\cdot;\theta_f); \theta_h\right)$, coordinates estimator $\phi(\cdot;\gamma)$
\State \textbf{Phase 1: prototype generating} 
\State Train $F(\cdot;\theta)$ by $\min\mathcal{L}_{task}(X, Y) = \min_{\theta_f, \theta_h} \mathcal{L}(G_h\left(G_f(X; \theta_f); \theta_h\right), Y)$ (Eq.2)
\State Obtain global prototypes $\mathcal{B} = \{\beta_k\}_{k=1}^K \in \mathbb{R}^{K\times d}$ through applying K-Means clustering to the output of the trained backbone $G_f(X; \theta_{f})$
\State \textbf{Phase 2: representation projecting} 
\State Re-initialize the parameters of $F(\cdot;\theta) = G_h\left(G_f(\cdot;\theta_f); \theta_h\right)$
\While{$\mathcal{B}, \theta_f, \theta_h, \gamma$ has not converged}
    \State Sample minibatch of size $n_b$ from $D$\;
    \For{$i \gets 1$ to $n_b$}
        \State Obtain $i$-th instance representation distribution by $P_i = \delta_{G_f(x_i;\theta_{f})}$ 
        \State Calculate $i$-th instance projection representation coordinates $r_i = \phi(G_f(x_i; \theta_f);\gamma)$
        \State Obtain $i$-th instance projection representation distribution: $Q_i = \sum_{k=1}^{K} r_{i}^{k} \delta_{\beta_{k}}$
        \State Calculate the OT distance between $P_i$ and $Q_i$: $\text{OT}(P_i, Q_i) = \min_{\textbf{T}_i\in \Pi (P_i, Q_i)} \langle \textbf{T}_i, \textbf{C}_i \rangle$
    \EndFor
    \State Average the OT distance within the minibatch to compute $\mathcal{L}_{projecting}(X, \mathcal{B}) = \frac{1}{n_b}\sum_{i=1}^{n_b} \text{OT}(P_i, Q_i) = \frac{1}{n_b}\sum_{i=1}^{n_b} \min_{\textbf{T}_i\in \Pi (P_i, Q_i)} \langle \textbf{T}_i, \textbf{C}_i \rangle$ (Eq.4)
    \State Compute $\mathcal{L}_{task}(X, Y) = \frac{1}{n_b}\sum_{i=1}^{n_b} \mathcal{L}(G_h(\sum_{k=1}^{K} r_i^k \beta_k;\theta_h), y_i)$ (Eq.5)
    \State Random select 50\% of the samples within minibatch to compute $\mathcal{L}_{diversifying}(X) = -\sum_{i=1}^{n_b}\sum_{j=1}^{n_b}\textbf{1}\{y_i, y_j \in \text{positive pair}\} \log{\frac{\exp{(\cos(r_i, r_j))}}{\sum_{i=1}^{n_b}\sum_{j=1}^{n_b} \exp{(\cos(r_i, r_j))}}}$ (Eq.6)
    \State Compute $\mathcal{L}_{orthogonalization}(\mathcal{B}) = (\frac{\|M\|_{1}}{\|M\|_{2}^{2}} + \max(0,|K-\|M\|_{1}|))$ (Eq.7)
    \State Update $\mathcal{B}, \theta_f, \theta_h, \gamma$ by minimizing $\mathcal{O} = \mathcal{L}_{task} + \mathcal{L}_{projecting} + \mathcal{L}_{diversifying} + \mathcal{L}_{orthogonalization}$ through gradient descent
\EndWhile
\State \Return $G_f, G_h, \phi, \mathcal{B}$
\end{algorithmic}
\end{algorithm}

\newpage
\subsection{Comparison with baseline deep networks.}
\label{appendix: experimental results}

\begin{table}[h]
\centering
\caption{Ablation results on the effects of different components in \method.}
\setlength{\tabcolsep}{2.5mm}{
\begin{tabular}{c|cccccc|c}
\toprule
                                                 & AD $\uparrow$  & HI $\uparrow$  & HE $\uparrow$  & JA $\uparrow$  & AL $\uparrow$  & CA $\downarrow$     & Win         \\ \midrule
MLP + \method                                    & \textbf{0.868} & \textbf{0.723} & \textbf{0.396} & \textbf{0.71}  & \textbf{0.964} & \textbf{0.489}      & 6           \\
\method w/o O                                    & 0.856          & 0.711          & 0.381          & 0.703          & 0.955          & 0.496               & 0           \\
\method w/o O, D                                 & 0.838          & 0.696          & 0.364          & 0.683          & 0.937          & 0.511               & 0           \\
w/o \method                                      & 0.825          & 0.681          & 0.352          & 0.672          & 0.917          & 0.518               & 0           \\ \hline
ResNet + \method                                 & \textbf{0.862} & \textbf{0.729} & \textbf{0.399} & \textbf{0.723} & \textbf{0.964} & 0.498               & 5           \\
\method w/o O                                    & 0.848          & 0.712          & 0.383          & 0.707          & 0.949          & \textbf{0.493}      & 1           \\
\method w/o O, D                                 & 0.83           & 0.698          & 0.368          & 0.69           & 0.933          & 0.518               & 0           \\
w/o \method                                      & 0.813          & 0.682          & 0.354          & 0.666          & 0.919          & 0.537               & 0           \\ \hline
FT-Transformer + \method                         & \textbf{0.871} & \textbf{0.738} & \textbf{0.397} & \textbf{0.738} & \textbf{0.97}  & \textbf{0.448}      & 6           \\
\method w/o O                                    & 0.859          & 0.722          & 0.383          & 0.725          & 0.96           & 0.453               & 0           \\
\method w/o O, D                                 & 0.841          & 0.704          & 0.369          & 0.702          & 0.94           & 0.466               & 0           \\
w/o \method                                      & 0.827          & 0.687          & 0.352          & 0.689          & 0.924          & 0.486               & 0           \\
\bottomrule
\end{tabular}}
\end{table}

\subsection{Computational efficiency details.}
\label{Appendix: computational efficiency details}
To approximate the optimal transport (OT) distance between two discrete distributions of size $n$, the time complexity bound scales as $n^2 \log(n) / \epsilon^2$ to reach $\epsilon$-accuracy with Sinkhorn’s algorithm, as demonstrated by ~\citep{chizat2020faster, altschuler2017near}. 
In this paper, for each instance $x_i$, we pushing $G_f(x_i;\theta_{f})$'s representation distribution $P_i$ to the corresponding projection representation distribution $Q_i$ by minimizing their OT distance. 
Thus the instance-wise time complexity bound scales as $O(K^2 \log(K) / \epsilon^2)$, where $K$ is set to the number of global prototypes.
The number of global prototypes $K$ is data-specific, and we set $K$ to the ceil of $\log(N)$, where $N$ is the total number of features.
While OT demonstrates strong computational capabilities for measuring the minimum distance between two distributions, it does not significantly increase computational complexity in our setting.

We also compare the computational cost of \method on a single GTX 3090 GPU. 
We report the computational cost (s) of \method per training epoch on different datasets.
As shown in Table ~\ref{appendix:computational cost table}, coupling deep models with \method produces a better performance on
all datasets with an acceptable cost.

\begin{table}[h]
\centering
\caption{Computational cost (s) per training epoch for \method.}
\label{appendix:computational cost table}
\begin{tabular}{ccccccc}
\toprule
                      & AD    & HI           & HE     & JA     & AL    & CA                    \\ \midrule
MLP + \method phase1  & 0.230 & 0.482        & 0.312  & 0.402  & 0.520  & 0.107                \\
MLP + \method phase2  & 1.030 & 2.435        & 0.909  & 1.204  & 1.469  & 0.543                \\ \hline
DCNV2 + \method phase1  & 0.436 & 1.000        & 0.653  & 0.840  & 1.081  & 0.210               \\
DCNV2 + \method phase2  & 1.217 & 2.831        & 1.159  & 1.485  & 1.934  & 0.599               \\ \hline
SNN + \method phase1  & 0.257 & 0.506        & 0.343  & 0.446  & 0.555  & 0.117                \\
SNN + \method phase2  & 1.025 & 2.371        & 0.926  & 1.162  & 1.528  & 0.533                \\ \hline
ResNet + \method phase1  & 0.413 & 0.919        & 0.616  & 0.783  & 1.019  & 0.196              \\
ResNet + \method phase2  & 1.131 & 2.697        & 1.167  & 1.451  & 1.908  & 0.568               \\ \hline
AutoInt + \method phase1  & 0.655 & 2.278        & 1.394  & 3.575  & 7.278  & 0.304              \\
AutoInt + \method phase2  & 2.009 & 9.976        & 5.578  & 22.884 & 23.444 & 0.707              \\ \hline
FT-Transformer + \method phase1  & 0.828 & 2.367        & 1.489  & 3.130  & 9.194  & 0.418        \\
FT-Transformer + \method phase2  & 1.579 & 4.176        & 1.976  & 3.798  & 9.962  & 0.804        \\
\bottomrule
\end{tabular}
\end{table}

\subsection{Sensitivity analysis.}
\label{Appendix: sensitivity analysis}

\textbf{The influence of loss weights.} 
Altogether, the proposed \method  aims to minimize the following objective function in stage 2:
\begin{equation}
\begin{aligned}
    \mathcal{O}= \mathcal{L}_{task}(X, Y) + \mathcal{L}_{projecting}(X, \mathcal{B}) + \mathcal{L}_{diversifying}(X) + \mathcal{L}_{orthogonalization}(\mathcal{B}).
\end{aligned}
\end{equation}
The weights of above losses are set to 1.0, 0.25, 0.25 and 0.25 respectively and the weights are fixed in all settings. 
We provide the sensitivity analysis for the loss weights conducted on FT-Transformer in Table ~\ref{appendix:sensitivity analysis}.
The results indicate that \method is robust to the loss weights. 

\begin{table}[h]
\centering
\caption{Sensitivity analysis for loss weights.}
\label{appendix:sensitivity analysis}
\begin{tabular}{ccccccc}
\toprule
Loss weights                   & AD $\uparrow$  & HI $\uparrow$  & HE $\uparrow$  & JA $\uparrow$  & AL $\uparrow$  & CA $\downarrow$       \\ \midrule
\textbf{1.0, 0.25, 0.25, 0.25}    & 0.871          & 0.738          & 0.397          & 0.738          & 0.97           & 0.448                 \\
1.0, 0.25, 0.25, 0.2             & 0.871          & 0.740          & 0.398          & 0.737          & 0.97           & 0.449                 \\  
1.0, 0.25, 0.25, 0.15             & 0.870          & 0.739          & 0.397          & 0.737          & 0.97           & 0.448                 \\ 
1.0, 0.25, 0.25, 0.1             & 0.869          & 0.737          & 0.395          & 0.736          & 0.969          & 0.450                 \\ 
1.0, 0.25, 0.25, 0.05             & 0.864          & 0.734          & 0.391          & 0.731          & 0.964          & 0.451                 \\ 
1.0, 0.25, 0.25, 0.0             & 0.859          & 0.722          & 0.383          & 0.725          & 0.96           & 0.453                 \\ \hline
\textbf{1.0, 0.25, 0.25, 0.25}    & 0.871          & 0.738          & 0.397          & 0.738          & 0.97           & 0.448                 \\
1.0, 0.25, 0.15, 0.25             & 0.869          & 0.737          & 0.397          & 0.736          & 0.969          & 0.451                 \\  
1.0, 0.25, 0.05, 0.25             & 0.869          & 0.737          & 0.395          & 0.734          & 0.969          & 0.452                 \\ 
1.0, 0.25, 0.0, 0.25             & 0.861          & 0.724          & 0.379          & 0.722          & 0.961          & 0.456                 \\
\bottomrule
\end{tabular}
\end{table}

\begin{minipage}{0.45\textwidth} 
\textbf{The influence of global prototype number to \method.}
Since the core of \method is the constructed P-Space, we further explore \method's performance under different global prototypes number.
Fig.~\ref{fig:prototype_number} shows the results of average performance of \method and its variants over all deep models. The \method’s performance gradually improves with global prototypes number increasing, and  finally reaches a stable level.
Thus it is reasonable to set the number of prototypes $K$ as the ceil of $\log(N)$, where $N$ is the total number of features. 
\end{minipage}
\hspace{0.05cm}
\begin{minipage}{0.53\textwidth}
\includegraphics[width=\linewidth]{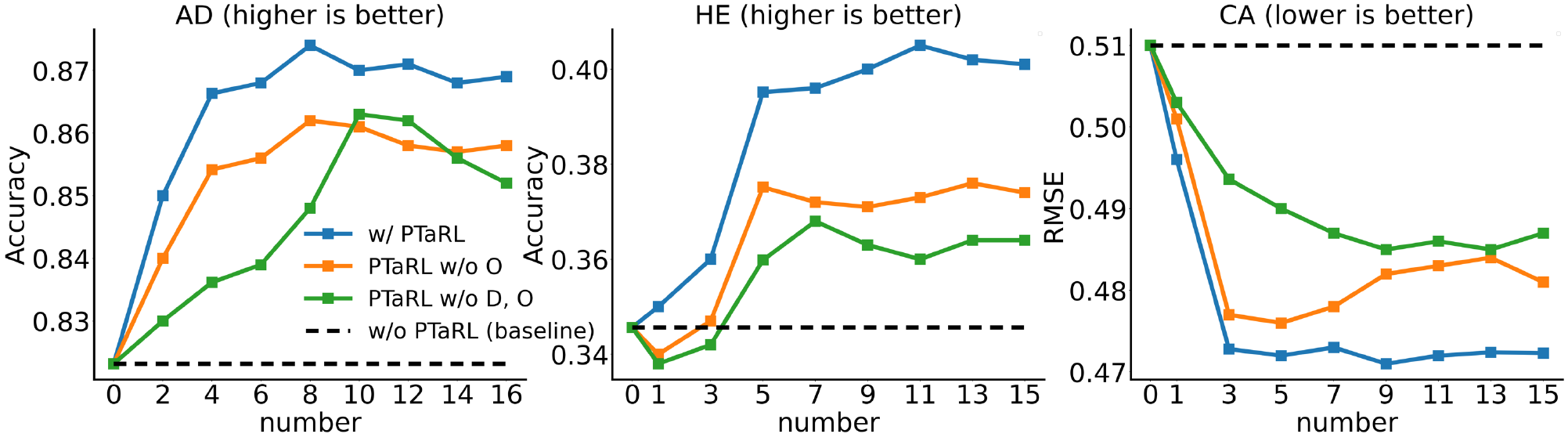}
\captionsetup{font=small}
\captionsetup{skip=15pt}
\captionof{figure}{\label{fig:prototype_number}Tabular prediction performance of the variants of \method with various numbers of prototypes.
The results are averaged over all baseline deep tabular models under binary classification (AD), multiclass classification (HE) and regression (CA).}
\end{minipage}%

\subsection{Additional results.}

\begin{table}[h]
\centering
\caption{The influence of DNN depths to the performance of \textsc{PTaRL}.}
\setlength{\tabcolsep}{2.5mm}{
\begin{tabular}{c|c|cccccc}
\toprule
Datasets & Models                                  & layers = 3     & layers = 5     & layers = 7     & layers = 9     \\ \midrule 
\multirow{4}{*}{HI $\uparrow$ }      & MLP         & 0.680          & 0.707          & 0.675          & 0.669          \\
& MLP + \textsc{PTaRL}                             & \textbf{0.719} & \textbf{0.730} & \textbf{0.729} & \textbf{0.732} \\
& FT-Transformer                                   & 0.687          & 0.719          & 0.709          & 0.691          \\
& FT-Transformer + \textsc{PTaRL}                  & \textbf{0.738} & \textbf{0.742} & \textbf{0.741} & \textbf{0.737} \\
\bottomrule
\toprule
\multirow{4}{*}{JA $\uparrow$ }      & MLP         & 0.670          & 0.715          & 0.704          & 0.689          \\
& MLP + \textsc{PTaRL}                             & \textbf{0.708} & \textbf{0.732} & \textbf{0.728} & \textbf{0.733} \\
& FT-Transformer                                   & 0.689          & 0.716          & 0.729          & 0.709          \\
& FT-Transformer + \textsc{PTaRL}                  & \textbf{0.738} & \textbf{0.741} & \textbf{0.745} & \textbf{0.742} \\
\bottomrule
\toprule
\multirow{4}{*}{CA $\downarrow$ }    & MLP         & 0.522          & 0.513          & 0.509          & 0.524          \\ 
& MLP + \textsc{PTaRL}                             & \textbf{0.491} & \textbf{0.480} & \textbf{0.482} & \textbf{0.479} \\
& FT-Transformer                                   & 0.486          & 0.476          & 0.472          & 0.478          \\
& FT-Transformer + \textsc{PTaRL}                  & \textbf{0.448} & \textbf{0.446} & \textbf{0.442} & \textbf{0.443} \\
\bottomrule
\end{tabular}}
\end{table}

\newpage
\subsection{Detailed explanation of the optimization process of Eq.~\ref{ot_loss}.}
\label{explanation of Eq. 4}
The Optimal Transport (OT) problem is usually to find the most cost-effective way to transform a given distribution to another distribution, which is typically achieved by calculating the specified transportation plan that minimizes the total transportation cost, while the minimized cost is usually called OT distance.
In our paper, we minimize the distance between original representation distribution over each sample $P_i$ by deep tabular models and the corresponding projection representation distribution $Q_i$ in P-Space with global prototypes, in order to preserve original data information (of $P_i$) in $Q_i$.
We follow the typical setting of OT problem to first estimate the transport plan to obtain the OT distance between $P_i$ and $Q_i$.
Then, the obtained OT distance is further used as loss function to jointly learn the two representations.

Specifically, after initializing the global prototypes $\mathcal{B}$ of P-Space, we project the original data samples into P-Space to learn the representations with global data structure information. 
To better illustrate the optimization process, we revise the Eq. 4 in the original paper to make it more readable. 
In Eq. 4, the $i$-th sample representation by deep tabular model is denoted as $G_f(x_i;\theta_f)$, the empirical distribution over this sample representation is $P_i = \delta_{G_f(x_i;\theta_f)}$, the projection representation distribution is denoted as: $Q_i = \sum_{k=1}^{K} r_{i}^{k} \delta_{\beta_{k}}$, where $r_i$ is coordinates. 
To capture the shared global data structure information, we formulate the representation projecting as the process of extracting instance-wise data information by $G_f(x_i;\theta_f)$ to $P_i$, and then pushing $P_i$ towards $Q_i$ to encourage each prototype $\beta_k$ to capture the shared global data structure information, a process achieved by minimizing the OT distance between $P_i$ and $Q_i$. 
The OT distance between $P_i$ and $Q_i$ could first be calculated by: $\text{OT}(P_i, Q_i) = \min_{\textbf{T}_i\in \Pi (P_i, Q_i)} \langle \textbf{T}_i, \textbf{C}_i \rangle$, where $C_{ik} = 1-\cos(G_f(x_i;\theta_f), \beta_k)$, the average OT distance between $P_i$ and $Q_i$ over train sets could be viewed as loss function $\mathcal{L}_{projecting}(X, \mathcal{B})$ to be further optimized: $\min\mathcal{L}_{projecting}(X, \mathcal{B}) = \min\frac{1}{n}\sum_{i=1}^{n} \text{OT}(P_i, Q_i) = \min_{\theta_{f}, \gamma, \mathcal{B}}  \frac{1}{n}\sum_{i=1}^{n} \min_{\textbf{T}_i\in \Pi (P_i, Q_i)} \langle \textbf{T}_i, \textbf{C}_i \rangle$, we use gradient descent to update $\theta_f, \gamma, \mathcal{B}$. 

\subsection{Detailed description of the \textit{global data structure information} and \textit{sample localization}.}
\label{explanation of global data structure}
In the context of any tabular dataset, we have observed \textit{global data structure information} comprises two different components: (i) the global feature structure and (ii) the global sample structure.

Considering the feature structure, traditional and deep tabular machine learning methods utilize all features or a subset of features as input, allowing them to model inherent interactions among features and thereby acquire a comprehensive global feature structure.
In addition, there also exists the sample structure given a tabular dataset. Traditional methods (e.g., boosted trees) can effectively model the overarching relationships between data samples. Specifically, in XGBoost, the dataset undergoes partitioning by comparing all the samples, with each node of a decision tree representing a specific partition, and each leaf node corresponding to a predictive value. The iterative splitting of nodes during training empowers decision trees in XGBoost to learn the distribution of all the samples across distinct regions of the data space, capturing global sample structure.

However, we note that deep tabular machine learning methods typically rely on batch training to obtain data representations within a batch. These methods do not explicitly consider the structure between samples within a batch. 
Furthermore, they fail to capture the global structure between samples across different batches. This limitation presents challenges in comprehensively capturing global data distribution information, consequently impeding overall performance.

Our methods rebuild the representation space with global prototypes (P-Space) in the first stage.
Then in the second stage, the original data representation by deep tabular machine learning methods is projected into P-Space to obtain projection representation with global prototypes. 
On the one hand, by minimizing the Optimal Transport distance between the two representations, we could represent each sample with global prototypes, and in the meanwhile encode the global feature structure in learning global prototypes, considering backbone models can inherently learn the interactions among features.
On the other hand, the global prototypes are learned by directly modeling all the data samples and thus the complex data distribution could be obtained by global prototypes to capture the global sample structure.  
Therefore, \textsc{PTaRL} is able to capture both the feature and sample structure information by prototype learning. Considering previous deep tabular machine learning methods can only acquire the representations limited by the batch training, we use the concept of \textit{sample localization} to encapsulate this limitation.




\end{document}